\documentclass{article}

\usepackage{PRIMEarxiv}

\usepackage[utf8]{inputenc} 
\usepackage[T1]{fontenc}    
\usepackage{hyperref}       
\usepackage{url}            
\usepackage{booktabs}       
\usepackage{amsfonts}       
\usepackage{nicefrac}       
\usepackage{microtype}      
\usepackage{lipsum}
\usepackage{fancyhdr}       
\usepackage{graphicx}       
\graphicspath{{media/}}     

\usepackage[table]{xcolor}
\usepackage{adjustbox}
\usepackage{multirow}
\usepackage{diagbox}

\title{Unveiling the Anomalies in an Ever-Changing World: A Benchmark for Pixel-Level Anomaly Detection in Continual Learning}

\author{
  Nikola Bugarin \\
  University of Padova \\
  \texttt{nikola.bugarin@studenti.unipd.it} \\
  \And 
  Jovana Bugaric \\
  University of Padova \\
  \texttt{jovana.bugaric@studenti.unipd.it} \\
  \And 
  Manuel Barusco \\
  University of Padova \\
  \texttt{manuel.barusco@phd.unipd.it} \\
  \And 
  Davide Dalle Pezze \\
  University of Padova \\
  \texttt{davide.dallepezze@unipd.it} \\
  \And 
  Gian Antonio Susto \\
  University of Padova \\
  \texttt{gianantonio.susto@unipd.it} \\
}

\begin{document}
\maketitle

\begin{abstract}
Anomaly Detection is a relevant problem in numerous real-world applications, especially when dealing with images. However, little attention has been paid to the issue of changes over time in the input data distribution, which may cause a significant decrease in performance. In this study, we investigate the problem of Pixel-Level Anomaly Detection in the Continual Learning setting, where new data arrives over time and the goal is to perform well on new and old data.
We implement several state-of-the-art techniques to solve the Anomaly Detection problem in the classic setting and adapt them to work in the Continual Learning setting.
To validate the approaches, we use a real-world dataset of images with pixel-based anomalies to provide a reliable benchmark and serve as a foundation for further advancements in the field.
We provide a comprehensive analysis, discussing which Anomaly Detection methods and which families of approaches seem more suitable for the Continual Learning setting. 
\end{abstract}


\section{Introduction}
\label{sec:intro}

\begin{figure*}[!thbp]
  \centering
  \includegraphics[width=\linewidth]{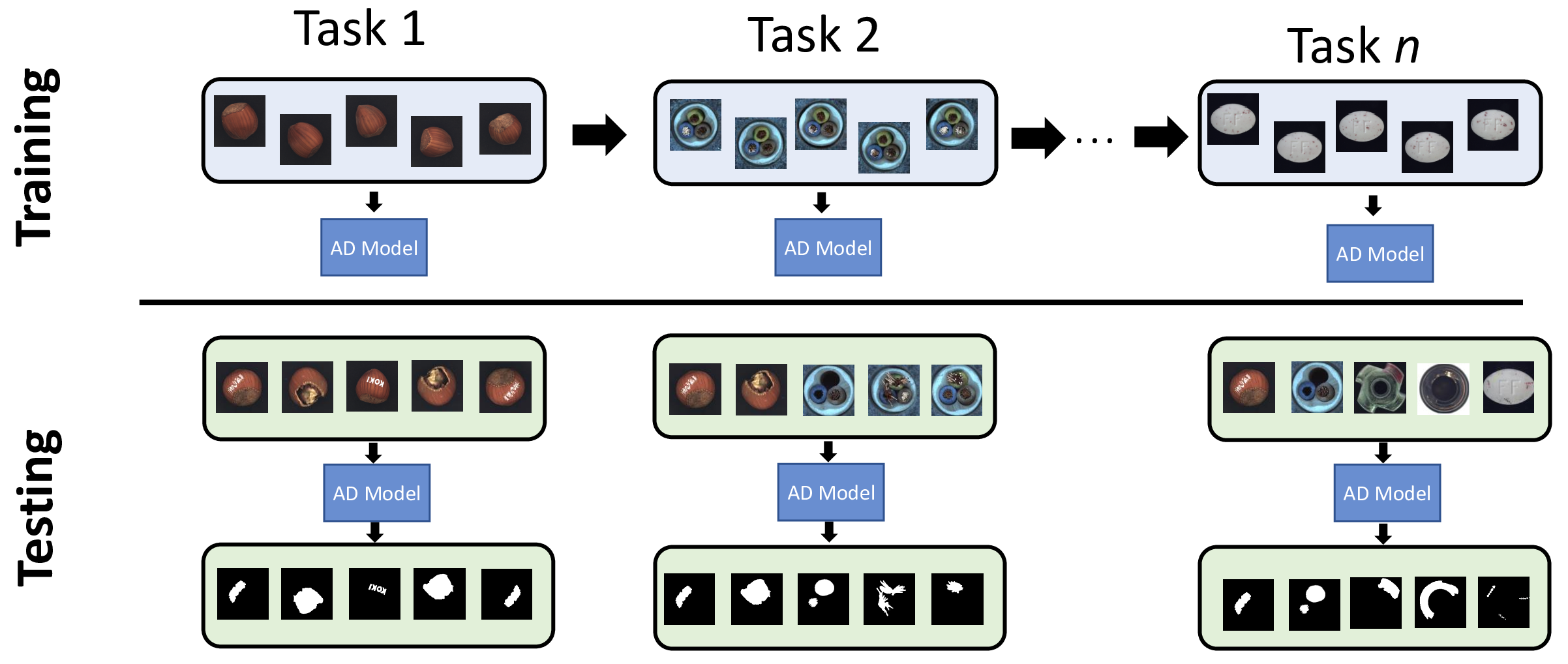}
   \caption{Considered CL setting for the AD problem. Each task corresponds to a new item. Our AD Model must be able to detect the anomalous products (image-level) and the defects inside the image (pixel-level) of a new item while remembering to perform well on previously seen items.}
   \label{fig:CLAD_scheme}
\end{figure*}

Anomaly Detection (AD) is an important and challenging problem in the fields of Machine Learning and Computer Vision.
Anomalies are patterns characterized by a noticeable deviation from the so-called normal data, where normal means compliant with some typical or expected features \cite{clvaead}.  
A significant advantage of the unsupervised techniques is that they do not require labeled data to learn from, making them easily deployable in many real-world applications.
Indeed, in many applications, avoiding label collection is crucial, considering that this process is time-expensive and resource-intensive, often requiring significant human effort and expertise.
This becomes even more pronounced when our focus lies on details at the pixel-level, which can incur considerably higher costs than at the image-level \cite{lin2014microsoft}.
Therefore, many approaches based on neural networks have been developed recently to address the problem of Unsupervised AD in the Computer Vision domain at the image-level and pixel-level \cite{bergmann2019,zavrtanik2021reconstruction,yu2021fastflow}.
\\
Despite these advancements, significant challenges still prevent these approaches from being deployed.
A significant issue is that shifts in the input data distribution can occur over time in real-world scenarios.
In these situations, the model must function successfully on older data while adjusting to the new input.
When these shifts appear repeatedly, we aim to train the model on a sequence of tasks, where each task represents a shift in the data.
For example, in the industrial sector, novel objects continually emerge, necessitating the identification of defects, while in the medical field, newly discovered body structures require analysis to detect anomalies.
However, the neural networks are prone to the effect known as Catastrophic Forgetting: when learning a new task, artificial neural networks frequently forget the previous ones \cite{vandevan2019}.
This aspect significantly hinders the adoption of these approaches in real-world scenarios. 
A new branch of Machine Learning has been introduced, known as Continual Learning (CL) \cite{de2021continual} to address this issue. 
It focuses on learning from a stream of tasks to adapt to the new incoming tasks while remembering the previous ones.
\\
Therefore, in this work, we study the use of Continual Learning strategies in the framework of Anomaly Detection on a data stream, focusing on Anomaly Localization, also known as Pixel-Level Anomaly Detection.
We implement and test several well-known methods for AD that are considered state-of-the-art in the field.
In particular, we adapt these AD techniques to work in the CL setting by employing the well-known CL strategy Replay \cite{Replaypaper,Replaymethods} when the methods allow it or perform ad hoc modifications of the original methods to let them work properly in the CL setting.
Therefore, we perform a broad analysis by considering at least one approach for each category of the AD approaches in order to identify if some categories are more prone to forgetting than others.
Then, we discuss which methods are more easily adaptable to the CL setting, which are more prone to forgetting, and which seem more fitting for the CL scenario.
To summarize, this paper presents the following contributions:
\begin{itemize}
    \item We investigate the most well-known and state-of-the-art approaches to solving Pixel-Level Anomaly Detection in the Continual Learning setting.
    \item We adapt these AD techniques to work in the CL setting by employing the well-known CL strategy Replay when the methods allow it or performing ad hoc modifications of the original methods to enable them to work well.
    \item We provide a comprehensive analysis, discussing which AD methods and which families of approaches seem more suitable for the CL setting. 
    
    

\end{itemize}

Moreover, to promote further research in the field and facilitate the rapid development of new methods along with comparisons with state-of-the-art approaches, we have made the code used available \footnote{https://github.com/dallepezze/clad}.
The paper is structured as follows: Sec. \ref{sec:related_work} provides an overview of the related work in the fields of CL and AD for Computer Vision. 
Sec. \ref{sec:considered_methods} details the AD methods implemented in our study and their adaptation for the CL setting. 
Sec. \ref{sec:experimental_setting} presents the Experimental Setting, which includes information on the metrics used for evaluation and the benchmark dataset employed in our experiments.
The Results of our experiments are presented in \ref{sec:results}, where we analyze and discuss the performance of the adapted AD methods in the CL setting.
Finally, in \ref{sec:conclusion_future_work}, we provide the Conclusion and discuss potential future research directions.


\section{Related Work}
\label{sec:related_work}
We delineate the relevant literature as follows:
Sec. \ref{subsec:intro_cl} presents the CL framework. 
In Sec. \ref{subsec:intro_ad_cv}, we outline the approaches for Unsupervised AD. Conversely, Sec. \ref{subsec:cl_for_ad} delves into the intersection of CL and AD, highlighting previous research in this area.

\subsection{Continual Learning}
\label{subsec:intro_cl}

In the traditional Machine Learning setting, models are trained on fixed datasets.
However, when considering real-world applications, it is easy to assume that the environments where the model is deployed could see new data in the future with a different data distribution than that observed in training.
CL addresses this limitation by allowing models to adapt and learn from new data without forgetting what they have already learned.
More in detail, CL allows DL models to update and expand their knowledge over time, with minimal computation and memory overhead \cite{de2021continual}.
Thus, an effective CL solution is expected to have low forgetting, require low memory consumption, and be computationally efficient.
\\
The methods in the CL literature can be grouped into three big families of techniques known as rehearsal-based \cite{shin2017continual,rolnick2019experience,chen2023online,atkinson2021pseudo,luo2023relay}, regularization-based \cite{pomponi2020efficient, ramapuram2020lifelong,han2022online}, and architecture-based approaches \cite{sokar2021spacenet, mallya2018piggyback,rajasegaran2019random}.
Rehearsal-based techniques assume storing and reusing past data samples during training.
Within this category, various approaches exist, but one of the most renowned methods is referred to as Experience Replay \cite{rolnick2019experience}, also known as Replay.
Regularization-based approaches consider additional constraints or penalties during training to maintain the memory of old tasks. 
For example, they could penalize the parameters differently based on the importance \cite{kirkpatrick2017overcoming} or perform distillation to maintain the knowledge learned in the previous tasks  \cite{li2017learning}.
Architecture-based approaches, as the name suggests, focus on changing the original model's architecture to help maintain the old knowledge, and the methods to perform this differentiate significantly among them \cite{rusu2016progressive, fernando2017pathnet, mallya2018packnet}.
\\
The related literature suggests that the Experience Replay approach appears to be the most effective and practical solution to reduce Catastrophic Forgetting \cite{Yangemp,pellegrini2020latent, buzzega2021rethinking,  kim2020imbalanced} when considering the image classification problem.
However, when delving into CL strategies, it is essential to recognize that what works for one problem may not seamlessly translate to another. 
For instance, while rehearsal-based methods have shown remarkable efficacy in addressing issues like image classification, their applicability in domains such as Object Detection results could be limited, and distillation-based approaches like Lwf prove to be more effective \cite{guan2018learn}.

Therefore, this highlights the importance of examining whether established methods, like Replay, are suitable for novel problems, such as Unsupervised AD.
Moreover, the AD approaches vary greatly among them (they are split into several categories and sub-categories).
This implies that a deep examination of how each AD technique performs based on a specific CL technique is necessary.
Therefore, in this work, we provide a deep analysis of the use of the Replay approach in the problem of Pixel-Level Anomaly Detection, examining which methods are more suitable to work in a stream scenario.
Moreover, as discussed below, while some AD techniques are well-suited to be used in the CL framework with the Replay approach, other methods are not adaptable to this technique. In these cases, we propose ad hoc modifications to let them operate in a continual scenario.

\subsection{Anomaly Detection}
\label{subsec:intro_ad_cv}

Unsupervised AD approaches find many applications in Computer Vision (CV), encompassing manufacturing, the medical domain, autonomous vehicles, security systems, and more.
In these contexts, detecting anomalies is crucial for safety, security, and efficiency. 
Many approaches proposed in the literature consider a model able to identify if a sample is anomalous.
This information is then provided to users to help them in their decision-making process.
However, they require more than just a binary outcome; they need an understanding of the reasons that brought the model to classify samples as anomalous.
Without this understanding, decision-makers may struggle to formulate appropriate responses or interventions.
Ensuring the interpretability of these systems can lead to safer and more efficient operations in various fields.
This is achieved by the capability to extend the model's predictions beyond image-level and into pixel-level granularity.
\\
The advantage of providing an interpretable mechanism is not the only key advantage.
Their unsupervised nature eliminates the need for a laborious label collection phase. 
This phase is typically time-consuming and resource-intensive, demanding substantial human effort and expertise.
\\
We can split most of the Pixel-Level AD approaches into two main families: reconstruction-based methods and Feature Embedding-based methods \cite{bao2023bmad, xie2024iad}.
\\
\textbf{Reconstruction-based methods} learn to reconstruct normal images during training. 
The idea is to use generative models for data reconstruction, where a large error during reconstruction indicates the presence of an anomaly.
This is the most historically significant research area, with approaches such as AutoEncoder and GANs.
\\
\textbf{Feature Embedding-based methods} consider data representations of images produced by a neural network, which is usually pre-trained. 
These approaches can be further categorized as Teacher-Student based, Normalizing Flow, and
Memory Bank. 
\\
\textbf{Teacher-Student approaches}, as the name suggests, are based on two networks, a teacher and a student architecture.
It exploits the knowledge distillation approach to transfer the learned knowledge, and when the features deviate, it is assumed the presence of an anomaly.
\\
\textbf{Memory Bank approaches} capture the features of normal images and store them in a feature
memory bank. Belonging to the category, three approaches are studied: Padim, PatchCore, and CFA (described with the rest of the methods in Sec. \ref{sec:considered_methods}).
All of these cannot be directly used for the Replay mechanism; therefore, ad hoc modifications are proposed to allow them to work effectively on a data stream.
\textbf{Normalizing Flow approaches}, as implied by the name, are based on the normalizing flow models to transform the complex input data distribution into a normal distribution. 
Then, it becomes possible to use the probability of the data under the distribution as a measure of its normality.

\begin{figure}[thbp]
  \centering
  \includegraphics[width=0.9\linewidth]{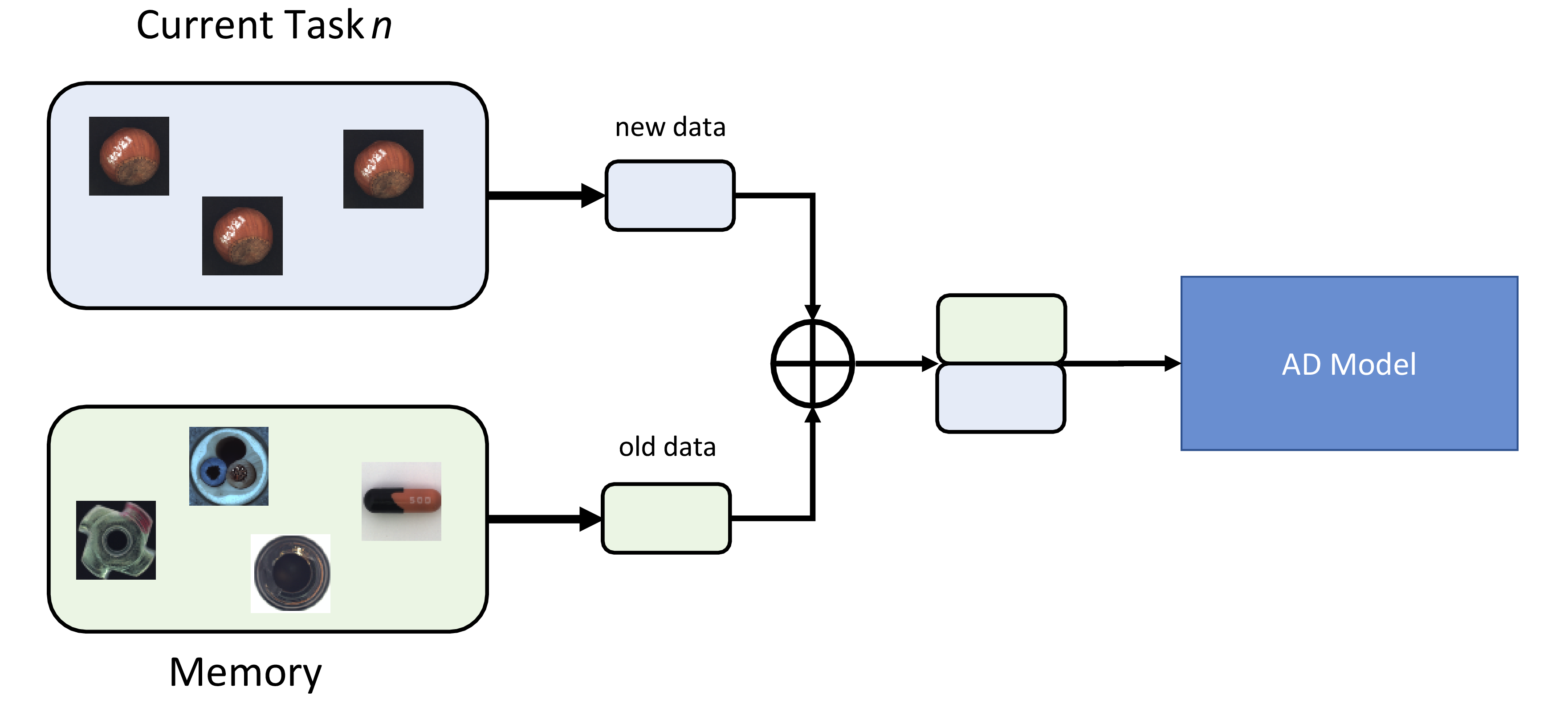}
   \caption{Scheme of the classic Replay approach employed to solve the Anomaly Detection problem in the Continual Learning setting.}
   \label{Fig:Replay_scheme}
\end{figure}

\subsection{Continual Learning and Anomaly Detection}
\label{subsec:cl_for_ad}
Even though AD is highly relevant in practice, only a few works in the literature address the problem of data distribution shifts for AD in the CL scenario.
Most of the existing works consider tabular data or time series.
For example, \cite{maschler2021regularization} studied the industrial domain, while \cite{hemati2021continual} considers the financial domain.
In the domain of Network Intrusion Detection, a series of works have been proposed in the literature \cite{clvaead}, \cite{amalapuram2022continual}, \cite{gonzalez2022steps}.
\\
When considering the CLAD topic (Continual Learning for Anomaly Detection) in the field of Computer Vision, few works have been conducted.
A study considering a sequence of tasks for AD was performed in \cite{frikha2021arcade}. In contrast to supervised and unsupervised paradigms, the authors propose a meta-learning approach. 
Nevertheless, it is crucial to note that their methodology exclusively operates at the image-level and involves treating an entire class as anomalous, resulting in a scenario that diverges significantly from realistic situations.

A series of works are proposed that work at image-level.
For example, \cite{gabbar2023incremental} consider X-ray computed tomography images for a nuclear power plant.
While they study and solve a realistic problem, their study exclusively examines image-level AD through supervised classification.
On the contrary, \cite{li2022towards} advocates for an unsupervised AD technique that retains samples from previous tasks. Despite moving from the supervised paradigm to the unsupervised one, this method exclusively operates at the image-level, diminishing its usefulness in real-world scenarios that demand interpretability.
A comprehensive exploration of various AD techniques was performed in \cite{xie2023iad} by assessing multiple paradigms, including the CL setting. 
The study compared several AD methods, demonstrating how employing these methods using the Fine-Tuning strategy (without CL solutions) leads to forgetting. Notably, the study does not present any solutions for adapting these methods to function effectively within the CL setting.
Finally, \cite{pezze2022continual} studies a series of AD techniques in the CL setting, but most of the approaches are classic approaches and far from the state-of-the-art in the field.
\\
Most of the work in the Computer Vision field focuses on predicting whether an image is normal or abnormal. However, Pixel-Level AD is frequently required in practice, given its several advantages, such as interpretability. 
Therefore, in our work, we implement several state-of-the-art AD techniques and make them work effectively in a data stream using the Replay approach when possible, or when the AD technique doesn't allow it, we propose some ad hoc modifications to make them work.


\section{Continual Learning Approach}
\label{subsec:replay_section}

Among the CL strategies proposed in the literature, the Replay approach appears to be the most effective and practical solution to reduce Catastrophic Forgetting \cite{Yangemp,pellegrini2020latent, buzzega2021rethinking} when considering the image classification problem.
Therefore, it is natural to consider it to be used to solve the AD problem in a data stream.
The Replay strategy consists of storing some randomly selected samples from the previous tasks in raw format and using them to maintain knowledge about previous tasks when training a new task. 
During training, each batch of data from the current task is combined with a batch of data from the memory of the previous tasks (sampled among them with the same probability).
The entire process of the Replay approach is shown in Fig. \ref{Fig:Replay_scheme}.
\\
It is crucial to note that this approach needs additional space for memorizing the subset of images from previous tasks. This implies that memory storage capacity is a major limitation; the model's performance decreases as memory capacity decreases.
Replay is a very generic approach that, as we will see, can be easily implemented for most of the AD techniques.
Therefore, unless differently reported, we implement the Replay approach for each AD technique as reported in Figure \ref{Fig:Replay_scheme}.
\\
However, this approach is unsuitable to work (entirely or partially) when considering the Memory Bank approaches: Padim, PatchCore, and CFA.
This is because these methods require a Memory Bank module that is not considered in the standard Replay approach. 
Therefore, for these approaches, specific ad hoc modifications are reported to let them work in a stream scenario.
In other words, when discussing reconstruction-based approaches (DRAEM), Student-Teacher approaches like STFPM and EfficientAD, and distribution map-based approaches like FastFlow, Replay is implemented without particular challenges.
Instead, when considering the Memory Bank approaches, ad hoc modifications are required as described below.

\section{Anomaly Detection Methods}
\label{sec:considered_methods}
We tested all the following AD methods, briefly describing their characteristics and how they were adapted to work in the CL setting.

\subsection{DRAEM}
\label{subsec:draem}
The Discriminatively Trained Reconstruction Embedding for Surface Anomaly Detection (DR{\AE}M) method \cite{draem} is a reconstruction-based method.
It is introduced to address a common drawback of generative AD methods. These methods only learn from anomaly-free data and lack optimization for discriminative AD since positive examples (anomalies) are unavailable during training. Training with synthetic anomalies leads to overfitting to synthetic appearances, resulting in poor generalization to real anomalies. To reduce overfitting, DRAEM proposes to train a discriminative model that considers the joint appearance of both reconstructed and original data, including the reconstruction subspace. Since it is purely based on learnable weights from neural network architectures, the classic Replay approach depicted in Fig. \ref{Fig:Replay_scheme} is employed here.

\subsection{STFPM}
\label{subsec:stfpm}
As the name suggests, Student-Teacher Feature Pyramid Matching (STFPM) \cite{st_pyramid} utilizes an architecture comprising two separate models for student and teacher, respectively.
In contrast to typical Knowledge Distillation methods, here the teacher and student networks in STFPM share the same architecture. 
The distinction is that the teacher network is pre-trained on Imagenet dataset, while the student network begins with random initialization. 
During the training phase, the objective is to enable the student network to reproduce the feature maps generated by the teacher. 
During the test phase, differences across different network levels are computed to produce the anomaly map of an image. 


\subsection{EfficientAD}
\label{subsec:efficientad}
EfficientAD \cite{batzner2024efficientad} employs a Student-Teacher methodology similar to  STFPM. 
However, it introduces a novel approach for feature extraction from a pre-trained neural network using a significantly reduced-depth model called Patch Description Network (PDN). 
Notably, Efficient AD utilizes a single teacher and a single student network based on the PDN architecture. 
Furthermore, to address logical anomalies such as misplaced objects, an autoencoder is integrated into the EfficientAD architecture. 
As for STPFM, the classic Replay approach is employed.

\subsection{Padim}
\label{subsec:padim}
Patch Distribution Modeling (PaDiM) is introduced by \cite{PaDiM}. 
Similar to other feature embedding-based techniques, it leverages a pre-trained CNN to extract patch embeddings. 
They propose that each position of a patch in the image can be characterized by a multivariate Gaussian distribution. 
During the inference process for a given test image, the Mahalanobis distance is computed for each patch, providing the anomaly score.
In this approach, there are no weights from a neural network to be updated, only a set of multivariate Gaussian distributions, one for each patch.
Typically, this implies considering a different set of Gaussian distributions for each task, but this violates constraints on the memory size that should not increase linearly with the number of tasks.
\\
Therefore, to allow the Padim approach to be updated with constant memory size, we propose an incremental averaging of Gaussian parameters when introducing new tasks (see the Supplementary Material \footnote{\label{note1}Supplementary Material can be found at https://github.com/dallepezze/clad} for more details).
The sum of old and new Gaussian distributions allows the memory to remain constant over time while allowing the model to have a representation of all tasks seen so far using a single Gaussian distribution per patch.

\begin{figure*}[thbp]
  \centering
  \includegraphics[width=\linewidth]{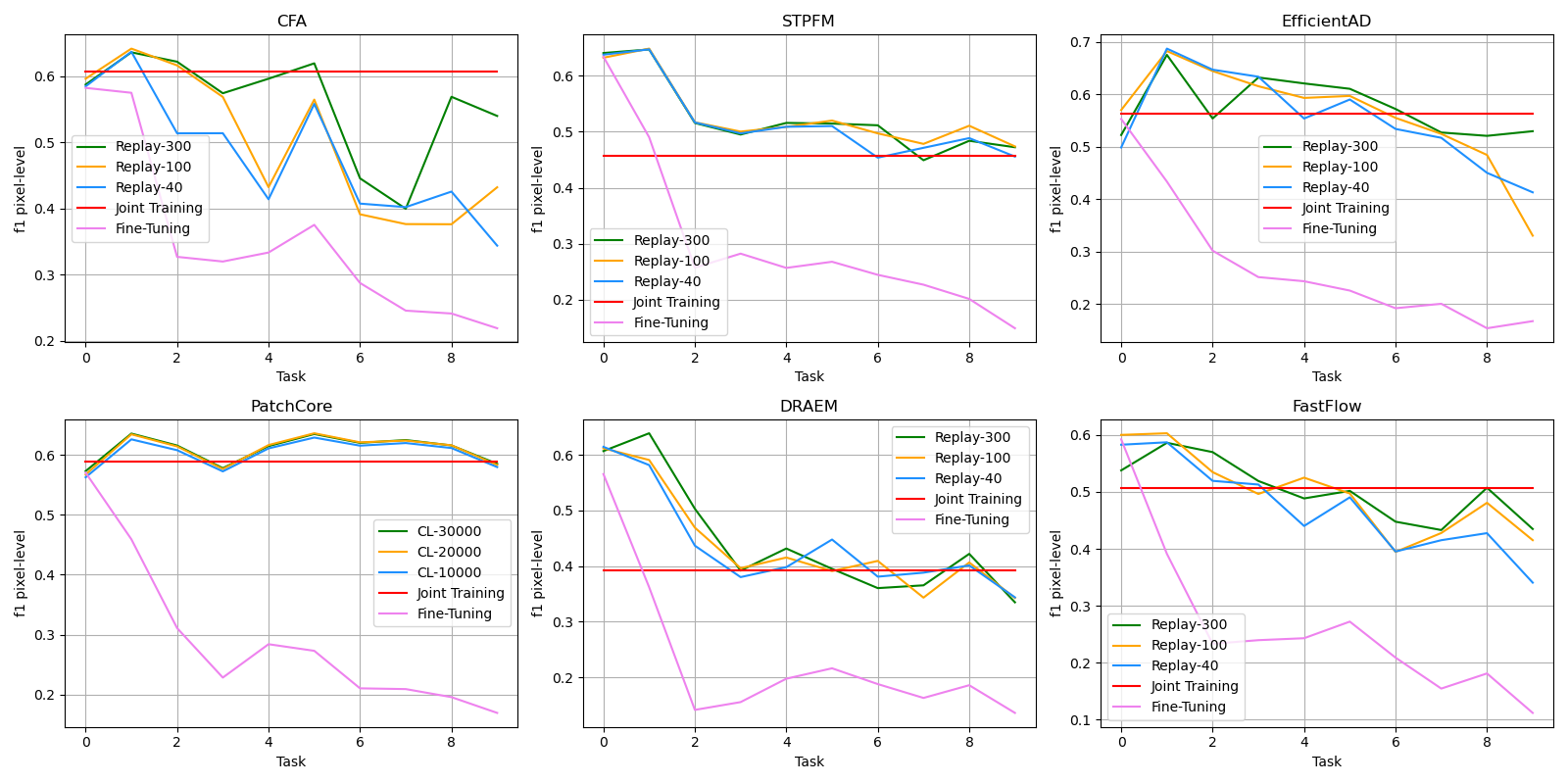}
   \caption{Each plot shows an AD technique tested in the CL setting. In detail, each plot shows the behavior of the f1 pixel-level metric over time, where each point represents the average performance on all the tasks seen so far. The considered strategies are the same as defined in Sec. \ref{sec:experimental_setting}.
   }
   \label{Fig:plot_f1_results}
\end{figure*}

\subsection{PatchCore}
\label{subsec:patchcore}
PatchCore \cite{patch} is a Memory Bank-based approach where patch features are stored within a memory bank.
However, to enhance efficiency, a coreset-reduction approach is employed to reduce the number of patches to be stored.
In this approach, there are no weights from a neural network to be updated, only the memory bank that needs to be addressed.

Upon the encounter of a new task, the coreset-reduction algorithm is utilized to identify $m$ patches to be retained, where $m=\frac{1}{N}memory~size$, with N denoting the number of tasks seen so far. 
The newly identified $m$ patches are incorporated into the memory bank. 
At the same time, the number of patches associated with the previous tasks is reduced to maintain only $m$ patches for each old task. 
To achieve this, the patches associated with each task undergo coreset-reduction until $m$ patches are attained. (More details in the Supplementary Material \footref{note1}  )

\subsection{CFA}
\label{subsec:cfa}
In \cite{lee2022cfa}, a novel Memory Bank approach called Coupled-Hypersphere-Based Feature Adaptation (CFA) is proposed.
Initially, a memory bank is created based on the features extracted from the training set.
Then, a trainable Patch Descriptor Network (PDN) is proposed with the aim of increasing the density of normal features (in alignment with the content stored in the memory bank).
The intuition is that this will allow us to distinguish the abnormal features more clearly.
\\
In this method, two distinct sources are identified that may contribute to the phenomenon of forgetting, the memory bank and the PDN network.
To address the memory bank problem, we propose to update the memorized patch features using an incremental average across batches of the current task training dataset  (further elaboration provided in the Supplementary Material).
After completing this initial procedure upon receiving the new task, we proceed to update the network parameters of the Patch Descriptor.
To perform the update of this part, we consider the Replay approach, consistent with its application in previous methods like DRAEM, STFPM, and EfficientAD.
The combination of these two mechanisms enables the model to learn from a stream of tasks while maintaining a constant memory size over time.

\subsection{FastFlow}
\label{subsec:fastflow}
FastFlow \cite{yu2021fastflow} utilizes 2D normalizing flows as a probability distribution estimator.
It takes as input the features obtained from a deep feature extractor.
During training, FastFlow learns to transform the input visual features into a tractable distribution, which is used to recognize anomalies in the inference phase.
This model employs the Replay approach as described in Sec. \ref{subsec:replay_section} and as represented in Fig. \ref{Fig:Replay_scheme}.

\section{Experimental Setting}
\label{sec:experimental_setting}
In the following part, we delineate the experimental setup employed in our conducted studies. Specifically, we outline the CL scenario in Sec. \ref{subsec:benchmark_dataset}. Following this, in Sec. \ref{subsec:metrics}, we explain the metrics under consideration, which are shown in Tab. \ref{tab:finalresults}. 
Lastly, we describe the methods present in Fig. \ref{Fig:plot_f1_results} used as comparison with the CL approaches.

\begin{table*}[th]
    \centering
    \caption{Performance comparison of Anomaly Detection strategies in the Continual Learning setting. Green reports the best value for each metric, while red shows the worst. All the metrics considered are described in Sec. \ref{sec:experimental_setting}, while each tested AD method is explained in \ref{sec:considered_methods}.}
    \label{tab:finalresults}
    \begin{adjustbox}{max width=1.2\textwidth,center}
    \begin{tabular}{|lc|c|c|c|c|c|c|c|}
    \hline
    \multicolumn{2}{|l|}{} & \multicolumn{1}{c|}{\textbf{\begin{tabular}[c]{@{}c@{}}Reconstruction \\ based\end{tabular}}} & \multicolumn{3}{c|}{\textbf{Memory Bank}} & \multicolumn{2}{c|}{\textbf{Student Teacher}} & \multicolumn{1}{c|}{\textbf{\begin{tabular}[c]{@{}c@{}}Normalizing \\ Flow\end{tabular}}} \\ \hline
    
    \multicolumn{2}{|c|}{\multirow{2}{*}{\textbf{\diagbox[width=13em]{Performance}{Strategy}}}} & 
    
    \multicolumn{1}{c|}{\textbf{DR{\AE}M}} & \multicolumn{1}{c|}{\textbf{PatchCore}} & \multicolumn{1}{c|}{\textbf{PaDiM}} & \multicolumn{1}{c|}{\textbf{CFA}}  & \multicolumn{1}{c|}{\textbf{STPFM}}  & \multicolumn{1}{c|}{\textbf{EfficientAD}}  & \multicolumn{1}{c|}{\textbf{FastFlow}} \\ \cline{3-9}
    
    \multicolumn{2}{|c|}{} & \multicolumn{1}{c|}{\textit{\begin{tabular}[c]{@{}c@{}}Replay\end{tabular}}} & \textit{\begin{tabular}[c]{@{}c@{}}CL\end{tabular}} & \multicolumn{1}{c|}{\textit{\begin{tabular}[c]{@{}c@{}}CL\end{tabular}}} & \multicolumn{1}{c|}{\textit{\begin{tabular}[c]{@{}c@{}}Replay\end{tabular}}} & \multicolumn{1}{c|}{\textit{\begin{tabular}[c]{@{}c@{}}Replay\end{tabular}}} & \textit{\begin{tabular}[c]{@{}c@{}}Replay\end{tabular}} & \textit{\begin{tabular}[c]{@{}c@{}}Replay\end{tabular}}\\ \hline
    
    \multicolumn{1}{|l|}{\multirow{2}{*}{\textbf{Image - level}}} & 
    \textit{AUC ROC} & 0.75 & {\cellcolor{green!25}0.97} & {\cellcolor{red!25}0.56} & 0.92 & 0.87 & 0.81 & 0.84 \\ \cline{2-9} \multicolumn{1}{|l|}{} & 
    \textit{f1}      & 0.87 & {\cellcolor{green!25}0.97} & {\cellcolor{red!25}0.83} & 0.94 & 0.89 & 0.89 & 0.89 \\ \hline 
    \multicolumn{1}{|l|}{\multirow{4}{*}{\textbf{Pixel - level}}} & 
    \textit{AUC ROC} & 0.82                      &  {\cellcolor{green!25}0.98}  & {\cellcolor{red!25}0.73} & 0.93 & 0.93 & 0.88 & 0.91 \\ \cline{2-9}  \multicolumn{1}{|l|}{} & 
    \textit{f1}      & 0.32                      &  {\cellcolor{green!25}0.58}  & {\cellcolor{red!25}0.17} & 0.53 & 0.48 & 0.49 & 0.42 \\ \cline{2-9}  \multicolumn{1}{|l|}{} & 
    \textit{PR AUC}  & {\cellcolor{red!25}0.11}  &  {\cellcolor{green!25}0.55}  & 0.29                     & 0.49 & 0.43 & 0.43 & 0.35 \\ \cline{2-9}  \multicolumn{1}{|l|}{} & 
    \textit{AU PRO}  & 0.65                      &  {\cellcolor{green!25}0.89}  & {\cellcolor{red!25}0.53} & 0.75 & 0.85 & 0.67 & 0.73 \\ \hline 
    
    \multicolumn{1}{|l|}{\multirow{1}{*} {\textbf{Time}}} & 
    \textit{training} & {\cellcolor{red!25}11h 46min} & 8min & {\cellcolor{green!25}6min} & 2h 16min & 27min & 5h 26min & 2h 19min  \\ \cline{1-9} 
    
    \multicolumn{2}{|l|}{\textbf{\begin{tabular}[l]{@{}l@{}}Architecture memory [MB] \end{tabular}}} & 
    389.6 & 275.6 & 275.6 & 301.2 & 93.6 & {\cellcolor{green!25}82.8} &  {\cellcolor{red!25}{440.80}} \\ \hline
    
    \multicolumn{2}{|l|}{\textbf{Additional memory [MB]}}  & 
    {\cellcolor{green!25}59.0} & 184.3 & {\cellcolor{red!25}1541.0} & 67.7 & {\cellcolor{green!25}59.0} &  {\cellcolor{green!25}59.0} & {\cellcolor{green!25}59.0} \\ \hline
    
    \multicolumn{2}{|l|}{\textbf{Relative gap ($\delta$) [\%]}} & 
    7.50 & {\cellcolor{green!25}0.01} & {\cellcolor{red!25}35.00} & 7.33 & 2.00 & 7.00 & 9.00 \\\hline
    
    \multicolumn{2}{|l|}{\textbf{Average forgetting [\%]}} & 
    {\cellcolor{red!25}27.08} & 0.72 & 19.12 & {\cellcolor{green!25}-4.70} & 14.78 & 19.11 & 17.39 \\ \hline

    \end{tabular}
    \end{adjustbox}
    \end{table*}

\subsection{Benchmark Dataset}
\label{subsec:benchmark_dataset}
The MVTec Dataset \cite{MVTec} is regarded as an extensive repository of images designed specifically for evaluating AD algorithms. This dataset encompasses ten objects and five textures, making it suitable for assessing the robustness and generalization capabilities of various AD techniques. 
In the given CL scenario, we consider a sequence of ten tasks, where each task corresponds to a different object (similarly as depicted in Fig. \ref{fig:CLAD_scheme}). Specifically, the evaluation includes the following objects: Bottle, Cable, Capsule, Hazelnut, Transistor, Metal Nut, Pill, Screw, Zipper, and Toothbrush.



\subsection{Metrics}
\label{subsec:metrics}

\subsubsection{Anomaly Detection metrics}
Various evaluation metrics are commonly employed to assess the performance of AD techniques using the MVTec dataset.
The first important categorization is based on how the evaluation is performed, whether on the image or pixel level.
For both of them, we consider the ROC and f1 metrics, as reported in Tab. \ref{tab:finalresults}.
For the pixel-level, we also consider the Per-region-overlap (PRO) metric, which ensures equal weighting of ground-truth regions regardless of their sizes, as explained in \cite{yang2020improving}. This stands in stark contrast to simplistic per-pixel metrics, where a single large correctly segmented region can compensate for numerous inaccurately segmented smaller ones.

\subsubsection{Continual Learning metrics}
The metrics described earlier are proposed within the CL setting. For each metric listed in Tab. \ref{tab:finalresults}, the final value is averaged across the set of tasks, as it is standard practice in CL evaluations, such as accuracy metrics.
Furthermore, the percentage of forgetting is reported to quantify the extent of forgetting across tasks. Additionally, to provide a more comprehensive assessment, the relative gap between the CL approach and the Joint Training strategy is highlighted. This comparison is essential as a model may exhibit low forgetting but substantially lower performance compared to the Join Training strategy, which is the upper bound for all the CL approaches.
\\
As previously stated, in the CL framework we are interested in updating and expanding their knowledge over time, avoiding forgetting, while at the same time we want to obtain this goal with minimal computation and memory overhead \cite{de2021continual}.
Hence, the values for memory and computation are also provided in Tab. \ref{tab:finalresults}.
For memory, we report the Architecture Memory and Additional Memory, indicating with Total Memory their combination.
With Additional Memory we refer to the memory necessary to store images for the Replay approach of patches for the Memory Bank based approaches.
Instead, Training Time is provided to assess computation.

\subsection{Considered CL Strategies}
During our work, we considered the comparison among the following CL approaches:
\begin{itemize}
    \item \textbf{Joint Training}: The model is trained on all tasks simultaneously, which is usually considered as an upper baseline for the CL strategies since they are not affected by the phenomenon of Catastrophic Forgetting.
    \item \textbf{Fine-tuning}: As a lower bound, we consider the fine-tuning approach, in which a model is presented sequentially only with data from the current task. 
    \item \textbf{Replay}: The approach as described in Sec. \ref{subsec:replay_section}.
    Specifically, we consider 300, 100, and 40 images that can be stored in the Replay memory, representing respectively 10\%, 5\%, and 2\% of the entire dataset.   
    \item \textbf{CL}: We report this generic signature for all the memory-bank approaches like Padim and PatchCore that do not fit the Replay category and have required ad hoc modifications to perform well in a data stream. In particular, for PatchCore, we vary the memory bank size to 30000, 20000, 10000.

\end{itemize}


\section{Results}
\label{sec:results}

This section analyzes the results outlined in Tab. \ref{tab:finalresults} and Fig. \ref{Fig:plot_f1_results}. 
For each AD method, a different plot is shown in Fig. \ref{Fig:plot_f1_results}, and a comprehensive comparison for each method and metric is also provided in Tab. \ref{tab:finalresults}.
We specifically examine the AD performance, memory consumption, and training time in Sec. \ref{subsec:results_AD_performance}, \ref{subsec:results_memory_performance}, and \ref{subsec:results_computation_performance} respectively.
Subsequently, in Sec. \ref{subsec:results_global} a generic analysis is provided for each method and AD families of techniques when considering all three factors together.

\subsection{Analysis on the performance metrics}
\label{subsec:results_AD_performance}
A general observation, as is notable from Fig. \ref{Fig:plot_f1_results}, is that for nearly all tested AD methods, the degree of forgetting is not particularly pronounced (except for the PaDiM approach). Moreover, the obtained performance resembles the ones of the upper bound Joint Training. 
Notably, the PatchCore method demonstrates no sign of forgetting, which is likely attributed to its reliance on a memory bank.
Of particular interest is the surprising performance of the CFA method, which appears to have improved instead of experiencing a decline in performance. 
This phenomenon could be attributed to the methodology of the CFA approach, which facilitates the gradual convergence of features from old and new tasks while effectively isolating anomalies.
\\
In contrast, all methods reliant on large networks with learnable parameters show some level of forgetting, typically falling within the range of 14.78\% to 19.11\%, 
except for DRAEM, which exhibits a higher forgetting rate of 27.08\%.
Notably, for most approaches, the gap concerning the upper bound is low and around 7\%.
Consequently, the Replay approach effectively addresses the AD problem within the CL framework.
In greater detail, the PatchCore method emerges as the optimal choice, achieving a f1 pixel-level score of 0.58. 
Good performance is also shown in CFA, STFPM, and Efficient AD (0.54, 0.49, 0.48, respectively). The lower performance is encountered with FastFlow (0.42) and DRAEM (0.32).



\subsection{Analysis on the memory consumption}
\label{subsec:results_memory_performance}
As discussed before, it is crucial to consider memory consumption when evaluating methods.
On this front, we can observe that PaDiM is the least efficient technique for analyzing memory consumption, consuming a total memory of 1816MB.
Following are the methods PatchCore, CFA, DRAEM, and FastFlow that have medium memory usage falling in the 369MB - 500MB range.
Then, the clear winners are the student teacher-based methods, i.e., EfficientAD and STFPM, which require only 142MB and 153MB, respectively.
These observations are made when considering the Total Memory, i.e., the combination of Architecture Memory and Additional Memory.
\\
Focusing solely on the Additional Memory component, which preserves data representations in methods like Replay or Memory Bank, can yield some interesting insights.
Despite the significant memory consumption, Padim continues to exhibit poor performance, utilizing 1541MB. 
In contrast, the method boasting the highest f1 pixel-level score, PatchCore, also demands the most memory, totaling 184.3MB. 
Notably, CFA follows with 67.7MB, followed by all Replay-based approaches (STFPM, EfficientAD, DRAEM, FastFlow), each requiring 59MB of Replay memory.
Essentially, Memory Bank-based approaches tend to require more memory than student-teacher-based, normalizing-flow-based, and reconstruction-based methods.




\subsection{Analysis on the Training Time}
\label{subsec:results_computation_performance}
When considering only the Training Time, Padim shows the fastest training time but consistently shows the poorest performance in terms of AD and Total Memory. 
Then Patchcore shows a training time of just 8 minutes. 
Another noteworthy method, STPFM, distinguishes itself with its rapid training duration of only 27 minutes. This contrasts with EfficientAD, another student-teacher-based approach, which requires considerably longer training time, totaling 5 hours and 26 minutes. Additionally, FastFlow and CFA demonstrate relatively good training times, each taking approximately 2 hours.
Conversely, DRAEM emerges as the slowest option, necessitating 11 hours and 46 minutes.

\subsection{Comprehensive Evaluation}
\label{subsec:results_global}
In this part, we provide a comparison, considering in conjunction all three critical factors: AD Performance, Memory, and Training Time.
The optimal choice depends on the specific requirements defined by the user.
For instance, we can assess PatchCore as the main winner when considering performance and training time.
However, other approaches could be adopted when including additional considerations, such as Total Memory, like the STFPM technique, which shows a good trade-off among all factors evaluated.
When considering the families of approaches in general, the Memory Bank-based approaches exhibit excellent AD performance and training efficiency but slightly higher memory consumption.
In contrast, student-teacher-based, normalizing flow-based, and reconstruction-based approaches seem to be good choices when the memory consumed is critical.
Nevertheless, the reconstruction-based approaches do not offer a good choice regarding AD and training time.





\section{Conclusions and Future Work}
\label{sec:conclusion_future_work}
In this work, we studied the Continual Learning Setting for the Pixel-Level Anomaly Detection problem.
This scenario is of paramount importance since traditional AD methods often operate under the assumption of stationary data distributions, 
which is not realistic in many practical scenarios.
In our study, we effectively integrate several state-of-the-art AD techniques into the CL framework. 
To make these AD techniques operate effectively in the CL setting, we propose using the well-known Replay approach to make them operate in a data stream or perform ad hoc modifications when Replay is not possible.
We report performance metrics for image-level and pixel-level anomalies to ensure a comprehensive evaluation using the well-known dataset MVTec.
In addition, we conduct a comprehensive exploration of various AD techniques within the CL framework.
In particular, we provide insights into each method's strengths and weaknesses, highlighting its performance, memory consumption, training time, and resilience to forgetting.
\\
Future research directions include the evaluation of additional AD techniques not considered in this study, which could provide a different insight concerning the approaches considered in this work.
As previously stated, we focused on rehearsal-based approaches to solve the AD problem in the CL setting.
Therefore, another direction is the investigation of other CL approaches that belong to families different from the rehearsal-based approaches.
Indeed, while Replay proved a viable and effective solution to solve the AD problem in the CL setting, we do not know the behavior of other approaches like distillation-based approaches (e.g., LwF) or methods like EWC.
Moreover, as pointed out in our study, future studies should focus on methods that try to optimize AD performance, memory, and training time simultaneously to guarantee deployment feasibility when considering resource-constrained environments.

\bibliographystyle{unsrt}  
\bibliography{references}

\end{document}